\documentclass[a4paper,fleqn]{cas-dc}
\usepackage{float}
\usepackage[authoryear,longnamesfirst]{natbib}
{}
{}
{}
\usepackage{graphicx}
\usepackage{algorithm}
\usepackage{algpseudocode}
\usepackage{caption}
\usepackage{subcaption}
\usepackage{amsmath}
\usepackage{amssymb}
\usepackage{eqparbox}
\newdimen{\algindent}
\algnewcommand\LeftComment[2]{%
	\hspace{#1\algindent}$\triangleright$ \eqparbox{COMMENT}{#2} \hfill %
}
\usepackage{mathtools, nccmath}
\DeclarePairedDelimiter{\nint}\lfloor\rceil

\usepackage{tikz}
\usetikzlibrary{shapes.geometric, arrows}

\tikzstyle{startstop} = [rectangle, rounded corners, 
minimum width=3cm, 
minimum height=1cm,
text centered, 
draw=black, 
fill=red!30]

\tikzstyle{io} = [trapezium, 
trapezium stretches=true, 
trapezium left angle=70, 
trapezium right angle=110, 
minimum width=3cm, 
minimum height=1cm, text centered, 
draw=black, fill=blue!30]

\tikzstyle{process} = [rectangle, 
minimum width=3cm, 
minimum height=1cm, 
text centered, 
text width=3cm, 
draw=black, 
fill=orange!30]

\tikzstyle{decision} = [diamond, 
minimum width=3cm, 
minimum height=1cm, 
text centered, 
draw=black, 
fill=green!30]
\tikzstyle{arrow} = [thick,->,>=stealth]

\def\tsc#1{\csdef{#1}{\textsc{\lowercase{#1}}\xspace}}
\tsc{WGM}
\tsc{QE}

\begin{document}
\let\WriteBookmarks\relax
\def\floatpagepagefraction{1}
\def\textpagefraction{.001}

\shorttitle{TrueDeep}    
\shortauthors{Pandey et al.} 

\title [mode = title]{TrueDeep: A systematic approach of crack detection with less data}  



%

\author[1]{Ramkrishna Pandey}



\ead{ramp@alum.iisc.ac.in}

\affiliation[1]{organization={GE Research},
            city={Bangalore},
            country={India}}

\author[1]{Akshit Achara}


\ead{f2016953p@alumni.bits-pilani.ac.in}



\begin{abstract}
Supervised and semi-supervised semantic segmentation algorithms require significant amount of annotated data to achieve a good performance. In many situations, the data is either not available or the annotation is expensive. The objective of this work is to show that by incorporating domain knowledge along with deep learning architectures, we can achieve similar performance with less data. We have used publicly available crack segmentation datasets and shown that selecting the input images using knowledge can significantly boost the performance of deep-learning based architectures. Our proposed approaches have many fold advantages such as low annotation and training cost, and less energy consumption. We have measured the performance of our algorithm quantitatively in terms of mean intersection over union (mIoU) and F score. Our algorithms, developed with 23\% of the overall data; have a similar performance on the test data and significantly better performance on multiple blind datasets. 
\end{abstract}

\begin{keywords}
 Cracks\sep Semantic Segmentation\sep Deep Learning\sep Computer Vision\sep Image Processing
\end{keywords}

\maketitle

\section{Introduction}\label{sec1}

Structural health monitoring is crucial for ensuring the safety and longevity of infrastructures such as roads, buildings, bridges, gas turbines, wind turbines, and pipelines. One of the key aspects of this monitoring is the detection and characterization of cracks, which can be a sign of potential failures in these structures. Traditionally, crack detection involves manual inspection of structures, which can be time-consuming, labor-intensive, expensive, and error-prone. With the rapid development of computer vision and machine learning algorithms, automatic crack detection systems have emerged as an efficient and reliable alternative. In this manuscript, we present a systematic approach of crack detection with less data using computer vision techniques, with a focus on segmentation-based methods. We discuss the challenges and opportunities in this field, provide a critical analysis of existing techniques, and highlight potential directions for future research. The ultimate goal of this manuscript is to provide a useful resource for researchers, engineers, and practitioners interested in automatic crack detection for structural health monitoring.

Image processing based techniques have emerged as one of the most effective methods for crack detection due to their ability to analyze large datasets quickly and accurately. Among these approaches, thresholding techniques have gained popularity as they enable the extraction of crack pixels from an image by setting a suitable threshold. Dynamic thresholding is often used to adjust the threshold value, making the algorithm adaptable to variations in image quality and crack size and shape.

The Canny edge detector algorithm [1] is widely used in image processing based crack detection due to its high accuracy in detecting edges. However, the algorithm's sensitivity to noise can cause issues with detecting edges in noisy images. As a result, noise reduction techniques are typically applied before the Canny algorithm to improve its performance. Despite these measures, challenges remain in using the Canny algorithm for crack detection, particularly when dealing with variations in image quality and texture [2].

Energy minimization methods, such as active contours, are commonly used for image segmentation tasks like contour detection, which can also be applied to crack detection. In [3], an initial estimate of crack regions was obtained from a crack saliency map. This map was generated by applying a steerable matched filter to the image, which enhanced the contrast between the cracks and the surrounding pavement. This initial estimate was fed into an active contour-based method, resulting in a better estimate of the cracks.

However, the performance of these methods depends on an initial estimate of the crack regions and the choice of multiple parameters such as regularization terms and focus parameters to converge towards the inside or outside of the region. The output of steerable matched filtering also depends on the choice of parameters such as filter size and scale tuning, which can be difficult to choose for varying images and complex features.

It should be noted that image processing based approaches depend heavily on the selection of appropriate parameters for each specific image. These parameters are typically selected based on factors such as image texture, color, brightness, scale, and noise levels. Proper selection of these parameters is essential to achieving accurate and reliable results.

Machine learning models like random forests [4], support vector machines [5], etc can be trained to identify crack and its characteristics using data, which involves handcrafting features and using these features for classifying into crack vs non-crack categories. These approaches are however sensitive to noise and different brightness levels[6].

Recently, deep learning based algorithms have gained popularity as they perform well on multiple tasks such as classification, segmentation, instance segmentation, object detection etc. Crack detection algorithms fall usually fall in one of these categories. One of the recent works in this direction which show promising results is a deep hierarchical CNN based approach (DeepCrack) for crack segmentation [7]. The method achieves an F-Score of 86.5 and outperforms the previous methods by a significant margin. The latest results on their source repository [8] show an improved F-Score of 86.84 using guided filters along with DeepCrack architecture (DeepCrack-GF).
In [9], the authors address a very relevant problem of dealing with partially accurate ground truths (specificially for thin cracks where the "All Black" phenomenon occurs) by using crack-patch-only (CPO) supervised
generative adversarial learning for end-to-end training to generate data. It compares the performance of crackGAN with several other CNN based models including DeepCrack[7] and mentions that the DeepCrack[7] training relies on accurate GTs and the method fails when the GTs are biased. In can be seen in the tables~\ref{table:F1_kaggle},~\ref{table:IOU_kaggle},~\ref{table:F1_best_kaggle} and~\ref{table:IOU_best_kaggle} that the performance of DeepCrack is better on thick cracks (crack500) as compared to thin cracks (cracktree200).
However, the above method requires large amount of training data to be able to generate images and the focus of the work is on thin cracks.

Overall, there is a scope of further improvements in the domain of crack detection that calls for further research. Many of the deep learning methods used for crack segmentation use localized parameters like dataset dependent thresholds which has some limitations on the generalization and automation.  The amount to data that is required to be annotation in order for reach sufficient model generalization also calls for some research.

In our work, we address the above issues by proposing a method that requires less data and annotations, along with several proposed augmentation approaches (discussed in section~\ref{augmentations}) for better generalization (see tables~\ref{table:F1_kaggle},~\ref{table:IOU_kaggle},~\ref{table:F1_best_kaggle} and~\ref{table:IOU_best_kaggle}).

\subsection{Contributions}\label{subsec1.1}

Following are our main contributions: 
\begin{itemize}
	\item Proposed an approach to select a subset of images from a given dataset to train a model that achieves a similar or better performance as compared to the same model trained using the entire dataset (see section~\ref{trueimageselection} for further details). 
	
	\item We have trained models with almost $\frac{1}{4}$ of the overall DeepCrack[5] data that achieve better or comparable results than the state of the art crack detection techniques on the same dataset. (see table~\ref{table:deepcrack})
	\item We have shown the effectiveness of our models by testing on multiple blind test datasets [10]. Our results are better or comparable to the state of the art techniques (trained on the same DeepCrack[7] dataset as our models) in the literature on these datasets [10]. See tables~\ref{table:F1_kaggle},~\ref{table:IOU_kaggle},~\ref{table:F1_best_kaggle} and~\ref{table:IOU_best_kaggle}. 
	\item We have also proposed data augmentation strategies(sw, sl, ss, mix) that improve the model performance on the blind datasets/unseen data. For more details, refer section~\ref{augmentations}.    
	
\end{itemize}

\section{Dataset}\label{dataset}

DeepCrack [7] dataset is used for all our experiments which has 300 train and 237 test images. This is a public benchmark dataset with cracks in multiple scales and backgrounds. The training images are of size $384\times544$ and the test images are of sizes $384\times544$  and $544\times384$. We used this dataset for building our models.

\subsection{Blind Test Dataset}\label{blinddataset}

The kaggle crack-segmentation-dataset [10] which contains the datasets listed in the Table~\ref{table:F1_kaggle} (except for the 'noncrack' and 'deepcrack' images that are present in the overall kaggle dataset). It is to be noted that we have used the test split of all these datasets. These datasets are used to blindly test the model performance and contain images of size $448\times 448$ with varying texture, color, crack characteristics, etc.

\subsection{TrueSet}
\label{trueset}

We followed a strategy discussed in the section~\ref{trueimageselection} to sub-sample 63 train and 7 validation images from the overall dataset [7] which contains 300 train images. The sub-sampled dataset is hereafter referred as trueset. The model obtained after training on trueset is hereafter referred as truecrack. The model trained on trueset is able to perform comparable with the state of the art methods in literature.

\subsubsection{True Image Selection}
\label{trueimageselection}

We have used prinical component analysis (PCA) to obtain our trueset (containing 70 images from 300 DeepCrack images) which approximates the distribution of overall DeepCrack data. We have used the encoder outputs obtained from a 2D UNet [11] with an EfficientNetB0 [12] backbone pretrained on imagenet [13]. PCA is applied on all the outputs to get the most significant dimension using the algorithm~\ref{alg:heltss}. We separate the images into different bins using hist function from matplotlib [14] and calculate the number of images to pick from each bin using the algorithm~\ref{alg:tspm}. We finally select the train and validation images from each bin with a 90-10 train-validation ratio using the algorithm ~\ref{alg:tss}. 

\begin{algorithm}
	\caption{Encoder Outputs to Coordinates (Coordinate Mapping): The algorithm returns a mapping of images and the corresponding first coordinate as obtained by PCA.}\label{alg:heltss}
	\begin{algorithmic}
		\Procedure{EncoderOutputs}{$images$}
		\State $outputs \gets \Phi$
		\For{$image$ in $images$}
		\State $output \gets encoder(image)$
		\State $output\_flat \gets convert\_to\_flat(output)$
		\Statex   \LeftComment{2}{map image to flat encoder output}
		\State $outputs \gets outputs \cup \{output\_flat\}$
		\EndFor
		\Statex   \LeftComment{1}{apply PCA to get the set of first dimension coordinates}
		\State $first\_coordinates \gets PCA(outputs, dims=1)$
		\Statex   \LeftComment{1}{$images$ is a set of all input images}
		\Statex   \LeftComment{1}{$images$ maps to $first\_coordinates$ (one-to-one mapping)}
		\State $coord\_map: images \rightarrow first\_coordinates$\\
		\Return $coord\_map$
		\EndProcedure
	\end{algorithmic}
\end{algorithm}

\begin{algorithm}
	\caption{Distance Based Selection: The algorithm sorts the images in order of its distances from the mean first coordinate and number of images to select from each bin.}\label{alg:tspm}
	\begin{algorithmic}
		\Procedure{GetSelectionParameters}{$coord\_map$}
		\Statex   \LeftComment{1}{$coord\_map : {images} \rightarrow {first\_coordinates}$  (algorithm~\ref{alg:heltss})}
		\State $num\_images \gets |{images}|$
		\Statex   \LeftComment{1}{average coordinate value of all the images}
		\State $avg \gets \frac {1}{num\_images}\sum_{i=1}^{num\_images} {first\_coordinates}$
		\State $distances \gets \Phi$
		\Statex   \LeftComment{1}{calculate the distance from $avg$ for each image}
		\ForAll{$coord \in {first\_coordinates}$}
		\State $distances \gets distances \cup \{\sqrt{(coord-avg)^2})\}$
		\EndFor
		\Statex   \LeftComment{1}{we have selected a value of 10 for our experiments}
		\State $n\_bins \gets 10$ 
		\Statex   \LeftComment{1} {$n$ is a list of count of the values in each bin}
		\Statex   \LeftComment{1} {$bins$ contains the edges of the bins. (Contains the lower}
		\Statex   \LeftComment{1} {and upper limit of each bin)}
		\State $n, bins \gets hist(distances, n\_bins)$ 
		\Statex \LeftComment{1}{$idx$ is a set of indices of bins}
		\State $idx \gets \{0, 1, 2, 3, 4, 5, 6, 7, 8, 9\}$
		\State $img\_bins \gets get\_images\_in\_each\_bin(bins)$
		\Statex \LeftComment{1}{A map of bin index and set of images in that bin}
		\State $bin\_mapping: idx \rightarrow {img\_bins}$
		\Statex \LeftComment{1}{A list of bin indices in descending order of number of}
		\Statex \LeftComment{1}{images in the bins}
		\State $idx\_descending \gets sort\_descending(n)$
		\Statex   \LeftComment{1}{selection parameter $\in [0, 1]$(0.5 was used for this paper)}
		\State $s \gets 0.5$
		\Statex \LeftComment{1}{Reduction fraction to reduce the images to select}
		\State $dec \gets \frac{s}{n\_bins-1}$
		\Statex \LeftComment{1}{A map of bin index and $|images|$ to be selected from that bin}
		\State $select\_mapping \gets \mathbb{R}^1 \mapsto \mathbb{R}^1$
		\State $select \gets \nint{\frac{num\_images \times s}{n\_bins - 1}}$
		\For{$bin_{idx} \in idx\_descending$}
		\If{$s > 0$}
		\State $select\_mapping[bin_{idx}] \gets select$
		\State $s \gets s - dec$
		\State $select \gets \nint{\frac{num\_images \times s}{n\_bins - 1}}$
		\EndIf
		\EndFor \\
		\Return $bin\_mapping, select\_mapping$
		\EndProcedure
	\end{algorithmic}
\end{algorithm}

\begin{algorithm}
	\caption{True Image Selection: Returns trueset which has a distribution similar to that of DeepCrack train dataset.}\label{alg:tss}
	\begin{algorithmic}
		\Procedure{SelectTrueImages}{$bin\_map, select\_map$}
		\Statex \LeftComment{1}{$\Phi$ refers to empty set}
		\State $images_{train}, images_{val} \gets \Phi$
		\For{$(b_{idx}, b_{img}) \in bin\_map$ }
		\State $cleaned \gets \Phi$
		\State $train \gets \Phi$
		\State $val \gets \Phi$
		
		\If{$b\_img = \Phi$}
		\State do nothing
		\Else 
		\State $select \gets select\_map[b_{idx}]$
		\If{$select > |b_{img}|$}
		\State $total \gets |b_{img}|$
		\Else
		\State $total \gets |select|$ 
		\EndIf
		\State $n_{train} \gets \nint{total\times 0.90)}$
		\State $n_{val} \gets (total - n_{train})$
		\If{$n_{val} > 0$}
		\State $jump \gets\nint{\frac{|b_{img}|}{n_{val}}}$
		\For{$z \gets 0$ to $n\_val$}
		\State $val_{idx} \gets \nint{z \times jump + \frac{jump}{2}}$
		\State $val \gets val \cup \{b_{img}[val_{idx}]\}$
		\State $cleaned \gets b_{img} - \{b_{img}[val_{idx}]\}$
		\EndFor
		\EndIf
		
		\If{$n_{train} > 0$}
		\State $jump \gets \nint{\frac{|cleaned|}{n_{train}}}$
		\Else
		\State $jump \gets 0$
		\EndIf
		
		\If{$jump = 1$}
		\State $train \gets train \cup {cleaned}$
		\ElsIf{$jump > 1$}
		\For{$z \gets 0$ to $n_{train}$}
		\State $train_{idx} \gets \nint{z \times jump + \frac{jump}{2}}$
		\State $train \gets train \cup \{cleaned[train_{idx}]\}$
		\EndFor
		\EndIf
		\State $images_{train} \gets images_{train} \cup train$
		\State $images_{val} \gets images_{val} \cup val$
		\EndIf
		\EndFor
		\State $trueset \gets images_{train} \cup images_{val}$\\
		\Return $trueset$
		\EndProcedure
	\end{algorithmic}
\end{algorithm}

\begin{figure}
	\centering
	\begin{tikzpicture}[node distance=1.3cm]
		
		\node (start) [startstop] {Start};
		\node (in1) [io, below of=start] {Input Images};
		\node (pro1) [process, left of=in1, xshift=-3cm] {Algorithm~\ref{alg:heltss}};
		\node (in2) [io, below of=pro1] {Coordinate Mapping};
		\node (pro2) [process, right of=in2, xshift=3cm] {Algorithm~\ref{alg:tspm}};
		\node (in3) [io, below of=pro2] {Select Mapping, Bin Mapping};
		\node (pro3) [process, left of=in3, xshift=-3cm] {Algorithm~\ref{alg:tss}};	
		\node (out1) [io, below of=pro3] {trueset};
		\node (stop) [startstop, below of=out1] {Stop};
		
		\draw [arrow] (start) -- (in1);
		\draw [arrow] (in1) -- (pro1);
		\draw [arrow] (pro1) -- (in2);
		\draw [arrow] (in2) -- (pro2);
		\draw [arrow] (pro2) -- (in3);
		\draw [arrow] (in3) -- (pro3);
		\draw [arrow] (pro3) -- (out1);	
		\draw [arrow] (out1) -- (stop);
		
	\end{tikzpicture}
	\caption{The flowchart shows the process of True Image Selection. The Input Images are the 300 DeepCrack train images used as an input for algorithm~\ref{alg:heltss} which results in a mapping of the image and its corresponding coordinate of the first dimension on applying PCA. The image-coordinate mapping is the input for algorithm~\ref{alg:tspm} which results in bins containing mutually exclusive image sets obtained based on the distance of an image from the mean distance over all images and the number of images to select from each of those bins. The algorithm~\ref{alg:tss} takes in these results from algorithm~\ref{alg:tspm} as inputs to produce a new train and validation set defined as trueset (see section~\ref{trueset}).} \label{fig:TSSflow}
\end{figure}
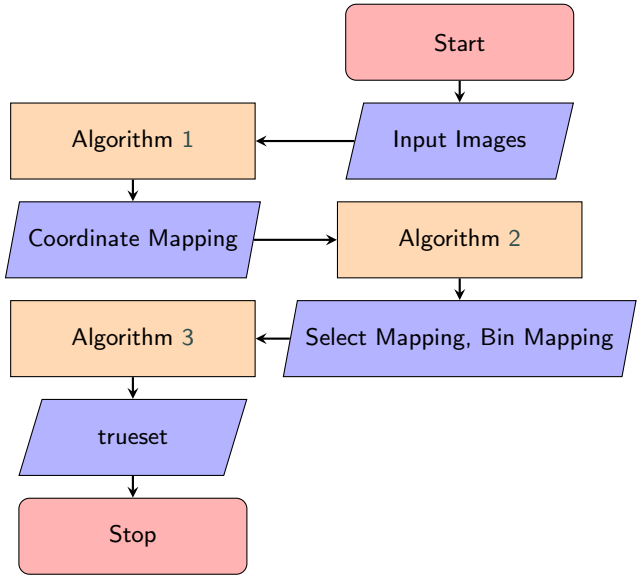

\begin{figure}[h]
	\centering
	\hfill
	\begin{subfigure}[t]{0.15\textwidth}
		\centering
		\includegraphics[width=\textwidth]{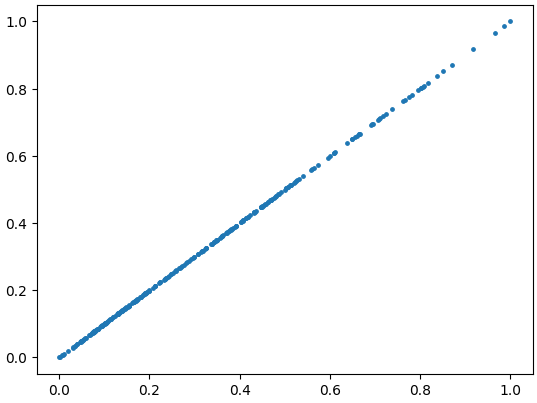}
		\caption{All the deepcrack train images.}
		\label{fig:overall1d}
	\end{subfigure}
	\hfill
	\begin{subfigure}[t]{0.15\textwidth}
		\centering
		\includegraphics[width=\textwidth]{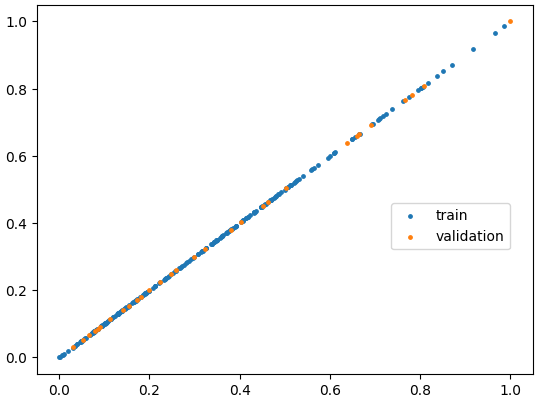}
		\caption{AllSet~\ref{allset}}
		\label{fig:allcrack1d}
	\end{subfigure}
	\hfill
	\begin{subfigure}[t]{0.15\textwidth}
		\centering
		\includegraphics[width=\textwidth]{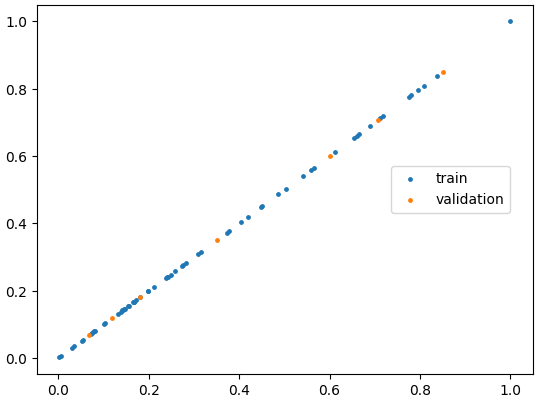}
		\caption{TrueSet~\ref{trueset}}
		\label{fig:truecrack1d}
	\end{subfigure}
	\caption{Shows the plots of first component of encoder outputs obtained after applying PCA. See algorithm~\ref{alg:heltss} for more details.}
	\label{fig:1d}
\end{figure}

\begin{figure}[h]
	\centering
	\hfill
	\begin{subfigure}[t]{0.15\textwidth}
		\centering
		\includegraphics[width=\textwidth]{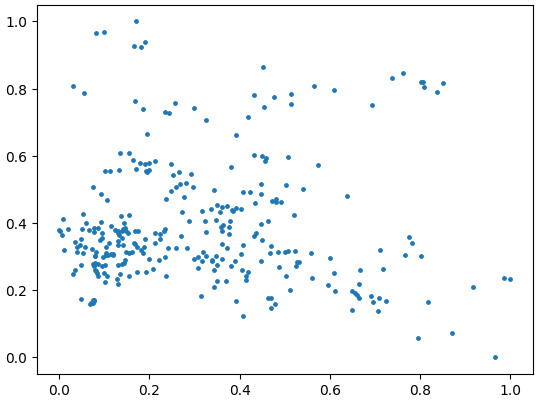}
		\caption{All the deepcrack train images. }
		\label{fig:overall2d}
	\end{subfigure}	
	\hfill
	\begin{subfigure}[t]{0.15\textwidth}
		\centering
		\includegraphics[width=\textwidth]{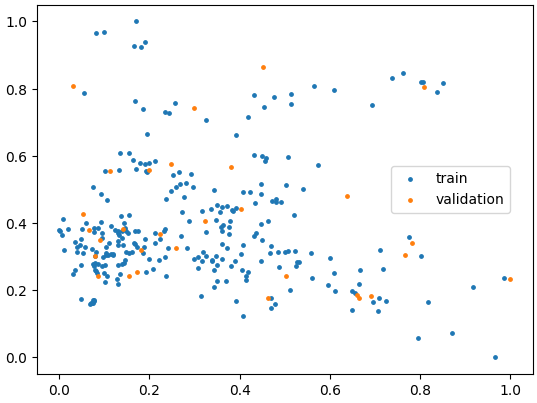}
		\caption{AllSet~\ref{allset}}
		\label{fig:allcrack2d}
	\end{subfigure}	
	\hfill
	\begin{subfigure}[t]{0.15\textwidth}
		\centering
		\includegraphics[width=\textwidth]{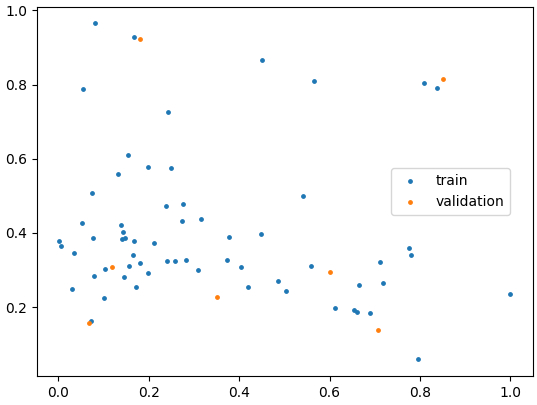}
		\caption{TrueSet~\ref{trueset}}
		\label{fig:truecrack2d}
	\end{subfigure}
	\caption{Shows the plots of first 2 components of encoder outputs obtained after applying PCA. }
	\label{fig:2d}		
\end{figure}

Figures~\ref{fig:1d} and~\ref{fig:2d} suggest that trueset follows similar distribution to the overall DeepCrack images visually where the former is a plot of images using the first component obtained from PCA and the latter is a plot of images using the first 2 components obtained from PCA.

\begin{figure}[h]
	\centering
	\includegraphics[width=0.40\textwidth,height=0.17\textheight]{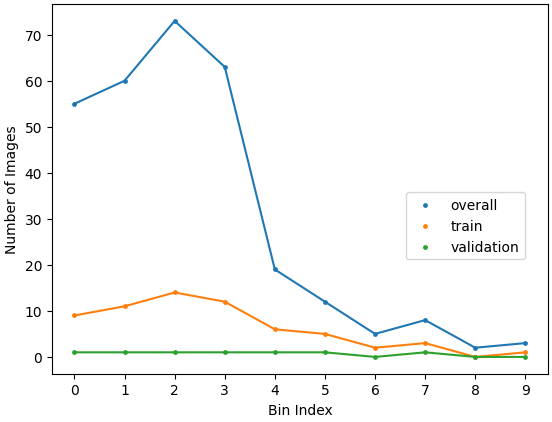}
	\caption{Figure shows the plot of the number of overall DeepCrack train images, trueset train images and trueset validation images present in each bin. The blue line shows the number of images present in each bin of the train split of the overall DeepCrack dataset (see section~\ref{dataset}); the orange line shows the number of trueset train images selected from each bin; the green line shows the number of trueset validation images selected from each bin. See section~\ref{trueset} and~\ref{trueimageselection} for more details.}
	\label{fig:binplot}
\end{figure}

It can be observed in Fig.~\ref{fig:binplot} that the number of  trueset train images and trueset validation images present in each bin are proportional to the number of overall DeepCrack train images. This proportionality demonstrates the similarity in the distribution of original training dataset i.e. DeepCrack train set and trueset.

\subsubsection{Feature Representation}
\label{feature representation}

The proposed true image selection approach uses the outputs from EfficientNetB0 backbone that is pretrained on imagenet, to obtain a feature represenation of the input images. Since the backbone was pretrained to learn feature representation of vast number of images with varying backgrounds and classes, the output feature representations capture both the local and global contexts from varying the input crack images with different local and global features. This representation is then used as an input to PCA to further process the representation as shown in the previous section. The proposed algorithms applied on the condensed outputs obtained from PCA result in separate groups of similar crack images. It is to be noted that there are other variants of EfficientNet which are more complex and give richer feature representations but since the task at hand was crack segmentation which doesn't include multiple objects, we have chosen a ligther variant.

\subsection{AllSet}
\label{allset}

We split the DeepCrack train images into 270 train and 30 validation images using the 90-10 ratio hereafter referred as allset.
The model trained on allset is hereafter referred as allcrack.

\subsection{Augmentations}
\label{augmentations}

We performed some augmentations on the trueset~\ref{trueset} (discussed in subsections~\ref{stochastic_width},~\ref{stochastic_length},~\ref{stochastic_scale_space}, and~\ref{stochastic_mix}) using domain knowledge before training the models. No augmentation was performed on the validation images.

\subsubsection{Stochastic width}\label{stochastic_width}
\begin{figure}[h]
	\centering
	\includegraphics[width=0.48\textwidth,height=0.10\textheight]{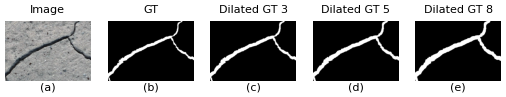}
	\caption{Shows (a) original image, (b) ground truth mask, (c) ground truth mask dilated with kernel of size 3$\times$3, (d) ground truth mask dilated with kernel of size 5$\times$5 and (e) ground truth mask dilated with kernel of size 8$\times$8.}
	\label{fig:stochasticwidth}
\end{figure}

Stochastic width augmentation is performed by dilating the ground truth using three different kernels of sizes $3\times 3$, $5\times 5$ and $8 \times 8$ respectively as shown in the figure~\ref{fig:stochasticwidth}. The augmented dataset contains $63$ (trueset images) + $3\times 63$(augmented images)$=252$ training images along with 7 validation images. For more details, refer [15]. The model trained on this dataset is hereafter referred as swcrack.

\subsubsection{Stochastic Length}
\label{stochastic_length}

\begin{algorithm}
	\caption{Random Masking}\label{alg:rm}
	\begin{algorithmic}
		\State C$ \gets list of connected components$
		\State $mask \gets ground\_truth$
		\ForAll{$c \in C$}
		\State $a \gets area(c)$
		\If{$a  \leq t_{0}$}
		\State do nothing
		\Else
		\Statex   \LeftComment{2}{Get the range of sqaure mask side}
		\State $range \gets getsiderange($c$)$
		\Statex   \LeftComment{2}{Estimate a polygon around the component $c$}
		\State $polygon \gets Polygon($c$)$
		\If{$a  \leq t_{1}$}
		\State $npts \in \{1, 2, 3\}$
		\ElsIf{$a  \leq t_{2}$}
		\State $npts \in \{2, 3, 4, 5\}$
		\Else
		\State $npts \in \{5, 6, 7, 8\}]$
		\EndIf
		\Statex   \LeftComment{2}{randomly get $npts$ within the polygon}
		\State $pts \gets rpwithinpolygon(polygon, npts)$
		\Statex   \LeftComment{2}{Remove the square mask pixels for each point}				
		\ForAll{$p \in pts$}
		\State $x \gets p.x$
		\State $y \gets p.y$
		\State $side \in [range.start\mathrel{{.}\,{.}}\nobreak range.end]$
		\State $gt\_mask \gets get\_square(side)$
		\State $mask \gets mask - gt\_mask$
		\EndFor
		\EndIf
		\EndFor
	\end{algorithmic}
\end{algorithm}

\begin{figure}[h]
	\centering
	\includegraphics[width=0.48\textwidth,height=0.15\textheight]{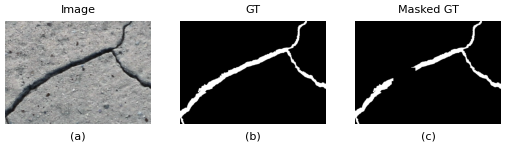}
	\caption{Shows (a) original image, (b) ground truth mask, and (c) randomly masked ground truth. }
	\label{fig:stochasticlength}
\end{figure}

In this approach, we perform the masking on ground truth as described in the algorithm~\ref{alg:rm}. We selected points (ranging from 1 to 3 for components of area between $t_0=50$ and $t_1=100$, 2 to 5 for compontents of area between $t_1$ and $t_2=200$ and 5 to 8 for components of area greater than $t_2$ ) randomly within each connected component in the ground truth and apply square masks with those points as the center of the squares. The augmentation results in two masks for each input image as shown in figure ~\ref{fig:stochasticlength}.
The augmented dataset contains $63$ (trueset images) + $63$ (augmented images)$=126$ training images and masks along with 7 validation images and masks.
The model trained using this dataset is hereafter referred as slcrack.

\subsubsection{Stochastic Scale Space}\label{stochastic_scale_space}

\begin{figure}[h]
	\centering
	\includegraphics[width=0.5\textwidth,height=0.12\textheight]{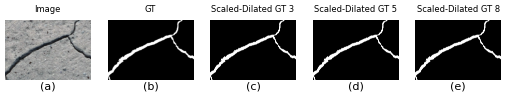}
	\caption{Shows (a) original image, (b) ground truth, (c) upscaled and dilated ground truth with a kernel of 3$\times$3, (d) upscaled and dilated ground truth with a kernel of 5$\times$5 and (e) upscaled and dilated ground truth with a kernel of 8$\times$8.}
	\label{fig:stochasticdifferentscale}
\end{figure}

\begin{figure}[h]
	\centering
	\includegraphics[width=0.5\textwidth,height=0.12\textheight]{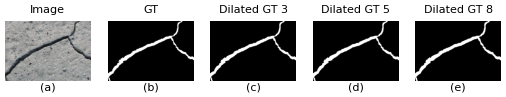}
	\caption{Shows (a) original image, (b) ground truth, (c) Scaled-Dilated GT 3 from fig.~\ref{fig:stochasticdifferentscale} downscaled to the same size as ground truth, (d) Scaled-Dilated GT 5 from fig.~\ref{fig:stochasticdifferentscale} downscaled to the same size ground truth and (e) Scaled-Dilated GT 8 from fig.~\ref{fig:stochasticdifferentscale} downscaled to the same size as ground truth. See section~\ref{stochastic_scale_space} for more details.}
	\label{fig:stochasticscalespace}
\end{figure}

We resized the ground truth mask 4 times the initial dimensions using the 'INTER\_CUBIC' [16] interpolation. Post that, we performed three dilation with kernels of sizes $3\times 3$, $5\times 5$ and $8 \times 8$ respectively as shown in the figure ~\Ref{fig:stochasticdifferentscale}. Then resized back each of the dilated images to the original dimension using the 'INTER\_NEAREST' [16] (nearest-neighbour) interpolation as shown in the figure~\Ref{fig:stochasticscalespace}. The dilation is applied in a different scale to preserve the sharpness in the crack boundaries. The method allows us to use a dilation kernel of fractional values on downscaling to the initial size of the image.

The augmented dataset contains $63$ (trueset images) + $3\times 63$ (augmented images)$=252$ training images and masks along with 7 validation images and masks. The model trained on this dataset will be referred as sscrack in the further sections.

\subsubsection{Combining length and width stochasticity: Mix}\label{stochastic_mix}

\begin{figure}[h]
	\centering
	\includegraphics[width=0.48\textwidth,height=0.10\textheight]{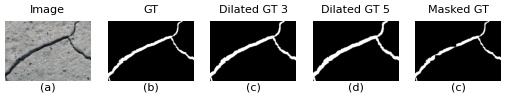}
	\caption{Shows (a) original image, (b) ground truth, (c) ground truth mask dilated with a kernel of size 3$\times$3, (d) ground truth mask dilated with a kernel of size 5$\times$5 and (e) randomly masked ground truth. }
	\label{fig:mixaugmentation}
\end{figure}

Based on our experiments, we have observed that the stochastic width method achieves high recall whereas stochastic length method high precision (see table~\ref{table:deepcrack} and ROC figure~\ref{fig:roc}). We have tried to balance the precision and recall by combining stochastic width and length approaches. We have obtained 3 augmentations from the input ground truth mask; the mask obtained after dilating the original mask with a kernel of sizes $3\times 3$ and $5\times 5$ respectively (for width stochasticity), and a randomly masked ground truth as discussed in section~\ref{stochastic_length} (for length stochasticity). The augmenation is shown in the figure ~\ref{fig:mixaugmentation}.

The augmented dataset contains $63$ (trueset images) + $3\times 63$(augmented images)$=252$ training images and masks along with 7 validation images and masks. The model trained on this dataset is hereafter referred as mixcrack.

\section{Training}\label{sec3}

We have used 2D UNet architecture [11] with a backbone, wherein weights are initialized with EfficientNetB0 [12] which is pretrained with imagenet [13] dataset.
\begin{figure}[h]
	\centering
	\includegraphics[width=0.4\textwidth,height=0.65\textwidth]{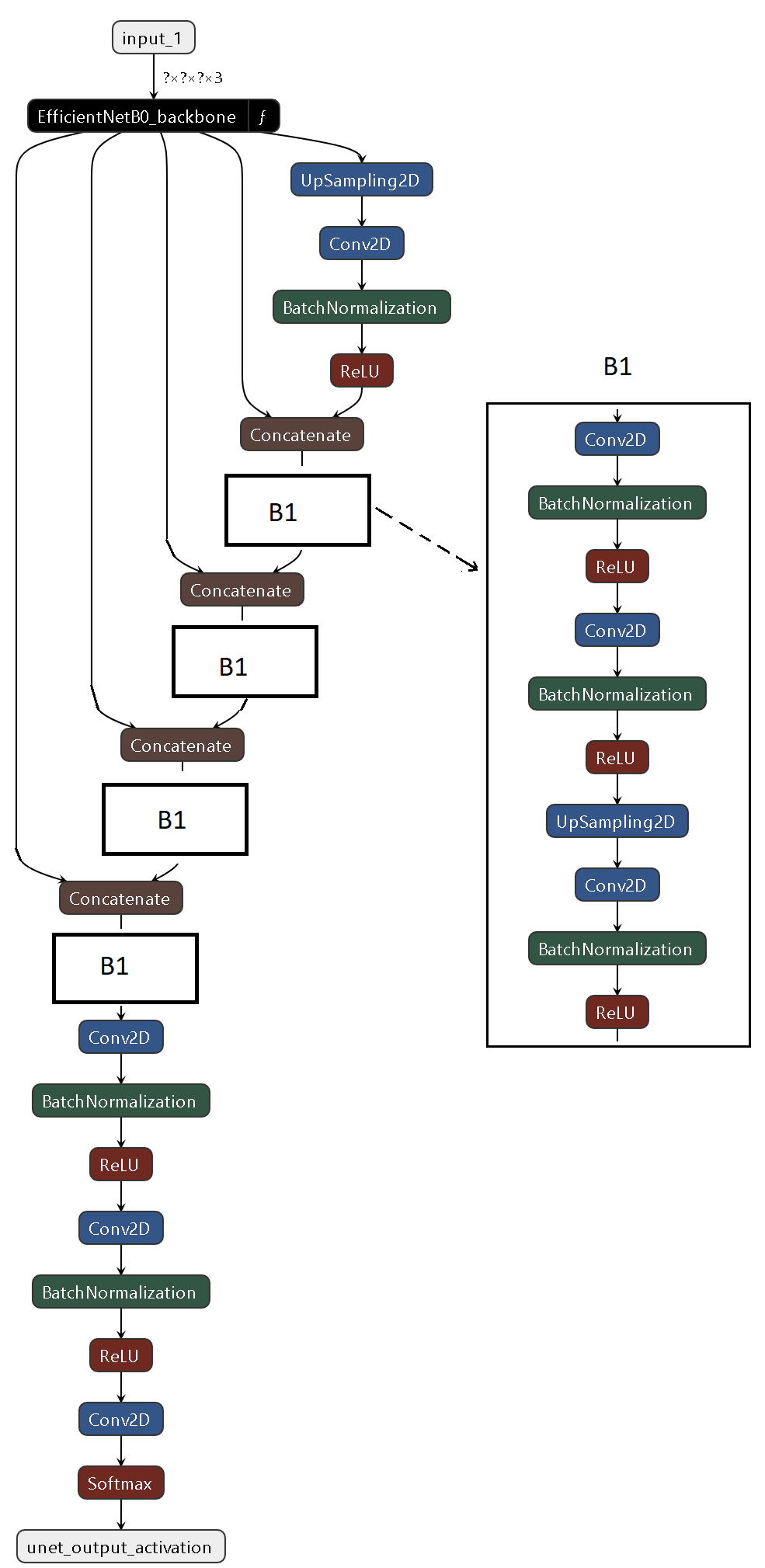}
	\caption{Shows architecture of the 2D UNet with EfficientNetB0 backbone used for all our experiments. (zoom to see details)}
	\label{fig:modelarch}
\end{figure}

The loss function used is binary focal dice loss (equation ~\ref{eq:3}) which is a sum of binary focal loss [17] (equation~\ref{eq:1}) and dice loss [18] (equation~\ref{eq:2}).

\begin{align}
	\centering
	L_{1}(G, P) = - G \alpha (1 - P)^\gamma \log(P) - (1 - G) \alpha P^\gamma \log(1 - P) 
	\label{eq:1}
\end{align}
\begin{align}
	\centering
	L_{2}(p , r) = 1 - (1 + \beta^2) \frac{p \cdot r}{\beta^2 \cdot p + r}
	\label{eq:2}
\end{align}
\begin{align}
	\centering
	L_{total} = L_{1}(G, P) + L_{2}(p, r)
	\label{eq:3}
\end{align} 

$G$, $P$, $\alpha$, and $\gamma$ are the ground truth, prediction, weighting factor and the parameter to decide the downweight amount respectively. Higher values of $\gamma$ indicates lower downweight. We have used $\alpha=0.5$ and $\gamma=3.33$ in our loss function. $p=\frac{TP}{TP+FP}$ and $r=\frac{TP}{TP+FN}$ are precision and recall respectively where $TP$, $FP$ and $FN$ are true positives, false postives and false negatives respectively. We have used $\beta=1$ in our experiments.

The batch size used for training is 8 and validation is 4.

The augmentations used during training are: flip, rotate, shift-scale-rotate, shear, translate, downscale, clahe, gaussian blur, median blur and sharpen using albumentations [19]. We randomly select one augmentation from the list of augmentations mentioned above to apply on each input image during the training process. Additional knowledge based augmentations performed before training are discussion in the section~\ref{augmentations}.

Initial learning rate of $1e-3$ is reduced using 'ReduceLROnPlateau' from keras [20] by a factor of 0.5 till $1e-6$ whenever the validation loss is not reducing for 50 continuous epochs.

\section{Experiments}\label{sec4}
\begin{figure}[h]
	\centering
	\includegraphics[width=0.3\textwidth,height=0.12\textwidth]{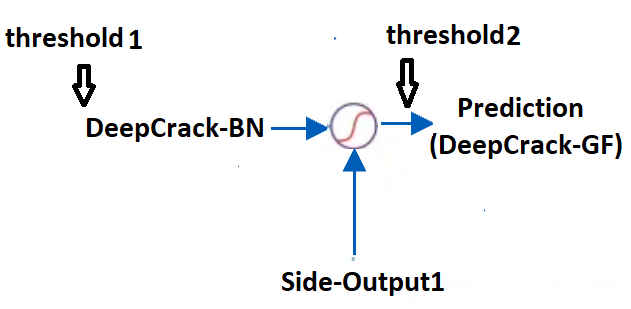}
	\caption{Show the process of obtaining the DeepCrack-GF output as discussed in [7].}
	\label{fig:dcgf}
\end{figure}

We evaluated all our models on the DeepCrack test dataset and blind test datasets (refer section~\ref{blinddataset}) without any preprocessing on images. We compared our results with DeepCrack-BN and DeepCrack-GF models (pretrained model downloaded from here [8]) using the testing scripts and framework provided in [8]. For more details on DeepCrack-GF, see Figure~\ref{fig:dcgf}
The evaluation metrics used are global accuracy (G), class average accuracy (C), mean intersection over union (mIoU) over all classes, Precision (P), Recall (R) and F-score (F) same as in [7]. Threshold is termed as T [7].

\section{Results and Discussions}\label{rds}

\subsection{Qualitative Analysis}\label{subsec5.1}
\begin{figure}[h]
	\begin{subfigure}[b]{0.48\textwidth}
		\centering
		\includegraphics[width=0.98\textwidth,height=0.11\textheight]{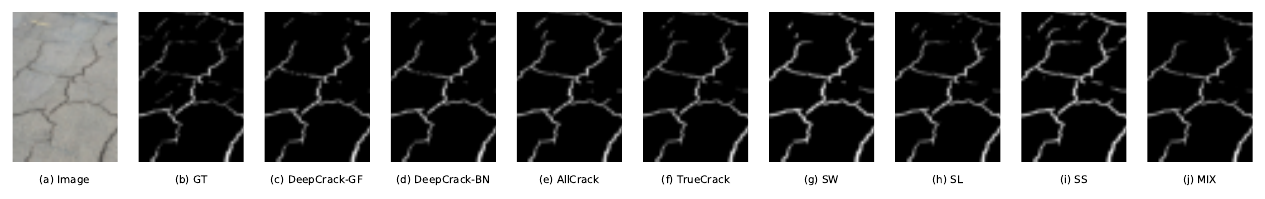}
		\caption{The results are calculated by applying a threshold on the predictions that results in the highest F score.}
		\label{fig:deepcrackinfbest1}
	\end{subfigure}
	\begin{subfigure}[b]{0.48\textwidth}
		\centering
		\includegraphics[width=0.98\textwidth,height=0.11\textheight]{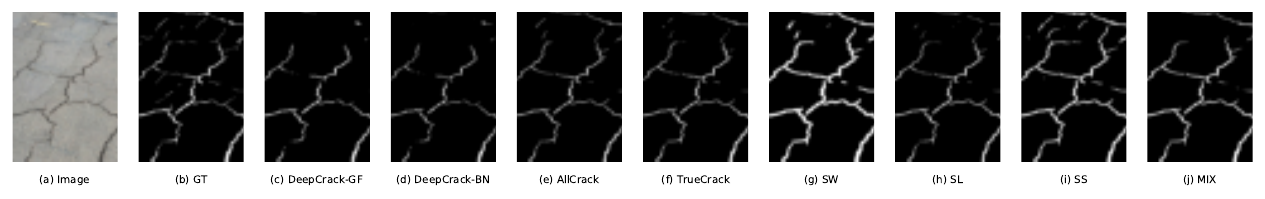}
		\caption{The results are calculated by applying a threshold of 0.5 on the predictions.}
		\label{fig:deepcrackinfstandard1}
	\end{subfigure}
	\caption{Figures show (a) original image, (b) ground truth, (c) DeepCrack-GF output, (d) DeepCrack-BN output, (e) allcrack output, (f) truecrack output, (g) swcrack output, (h) slcrack output, (i) sscrack output and (j) mixcrack output on the DeepCrack [7] dataset. Zoom to see the details.}
	\label{fig:deepcrackinf}
\end{figure}

\begin{figure}
	\begin{subfigure}[b]{0.48\textwidth}
		\centering
		\includegraphics[width=0.98\textwidth,height=0.11\textheight]{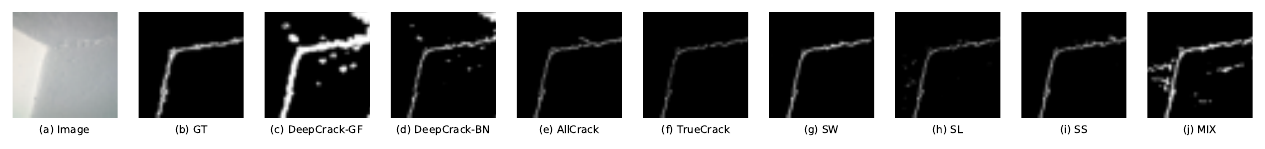}
		\caption{The results are calculated by applying a threshold on the predictions that results in the highest F score (see table~\ref{table:F1_best_kaggle}).}
		\label{fig:volkerbest1}
	\end{subfigure}
	\begin{subfigure}[b]{0.48\textwidth}
		\centering
		\includegraphics[width=0.98\textwidth,height=0.11\textheight]{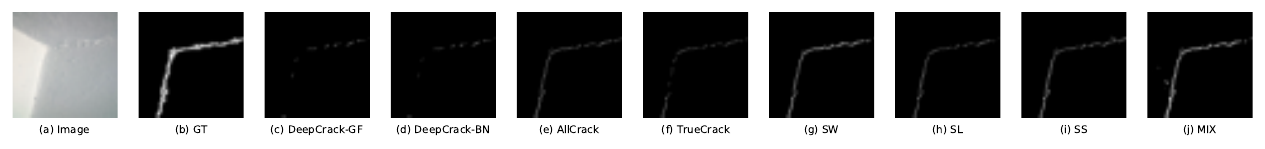}
		\caption{The results are calculated by applying a threshold of 0.5 on the predictions.}
		\label{fig:volkerstandard1}
	\end{subfigure}
	\begin{subfigure}[b]{0.48\textwidth}
		\centering
		\includegraphics[width=0.98\textwidth,height=0.11\textheight]{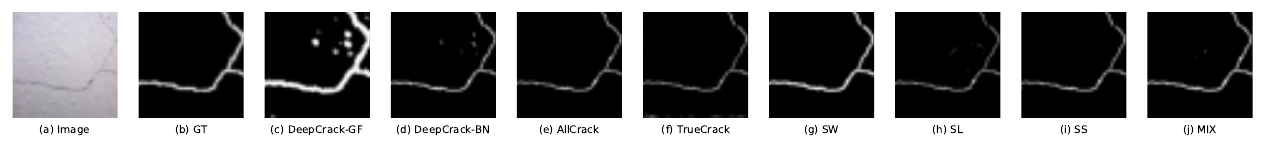}
		\caption{The results are calculated by applying a threshold on the predictions that results in the highest F score (see table~\ref{table:F1_best_kaggle}).}
		\label{fig:volkerbest2}
	\end{subfigure}
	
	\begin{subfigure}[b]{0.48\textwidth}
		\centering
		\includegraphics[width=0.98\textwidth,height=0.11\textheight]{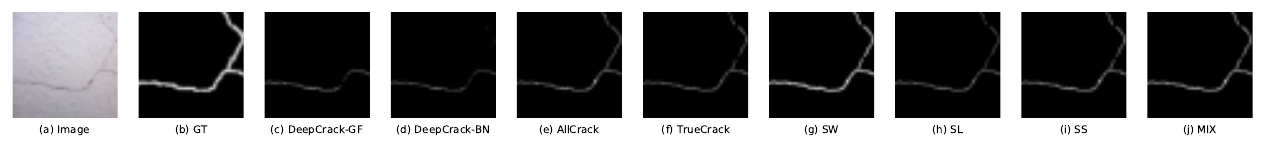}
		\caption{The results are calculated by applying a threshold of 0.5 on the predictions.}
		\label{fig:volkerstandard2}
	\end{subfigure}
	\caption{Figures show (a) original image, (b) ground truth, (c) DeepCrack-GF output, (d) DeepCrack-BN output, (e) allcrack output, (f) truecrack output, (g) swcrack output, (h) slcrack output, (i) sscrack output and (j) mixcrack output on the kaggle crack-segmentation dataset (Volker). }
	\label{volkerall}
\end{figure}

Fig.~\ref{fig:deepcrackinfbest1} is obtained at the best thresholds and fig.~\ref{fig:deepcrackinfstandard1} is obtained at a threshold of 0.5, show comparison of our results (e to j) with DeepCrack-BN and DeepCrack-GF approaches (c and d) on the DeepCrack dataset. It can be observed that our approaches have better predictions of cracks (zoom to see the finer details (top portion) of figure~\ref{fig:deepcrackinfbest1}, cracks are detected better in our techniques).
Figures~\ref{fig:volkerbest1},~\ref{fig:volkerstandard1},~\ref{fig:volkerbest2} and~\ref{fig:volkerstandard2} show comparison of our results (e to j) with DeepCrack-BN and DeepCrack-GF approaches (c and d) on a blind dataset(see section~\ref{blinddataset}).
Our techniques are more robust to background variations and have better crack detectability at both thresholds whereas DeepCrack results are having noisy predictions (when using the best thresholds) and are missing(when using a standard threshold of 0.5).

\subsection{ROC and PR Curves}\label{subsec5.2}

\begin{figure}[h]
	\begin{subfigure}[b]{0.48\textwidth}
		\centering
		\includegraphics[width=\textwidth, height=0.3\textheight]{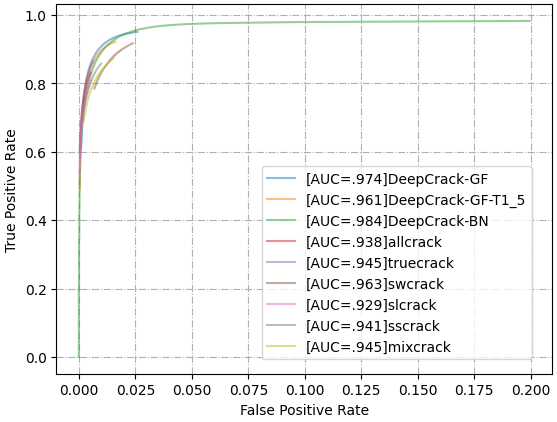}
		\caption{DeepCrack-BN has the highest AUC value of 0.984. It is to be noted that swcrack has a better AUC than the DeepCrack-GF-T1\_5 method. The scale on x-axis is 0-0.2 for clear visibility.}
		\label{fig:rocall}
	\end{subfigure}
	\hfill
	\begin{subfigure}[b]{0.48\textwidth}
		\centering
		\includegraphics[width=\textwidth, height=0.3\textheight]{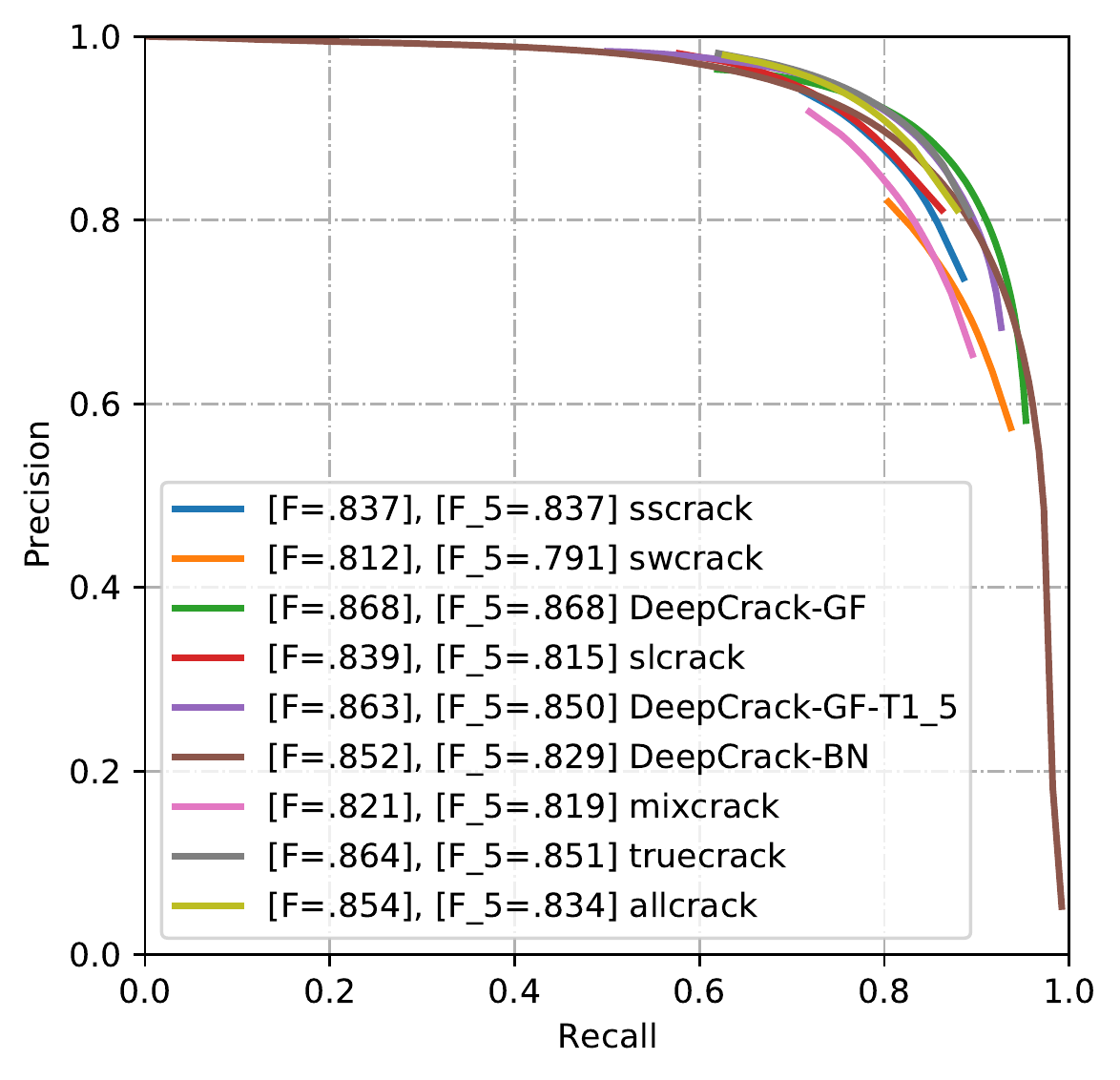}
		\caption{Precision-Recall curve comparison of our techniques with DeepCrack approaches on the DeepCrack dataset. F value is obtained at the threshold that results in the highest F-score and F\_5 is obtained at the threshold of 0.5. The
			truecrack approach achieves a F-score of 85.1 which is comparable to DeepCrack-GF-T1\_5. }
		\label{fig:prc}
	\end{subfigure}
	\caption{ROC and PR curves on the DeepCrack dataset. We have used a threshold1 value of 0.5 (see figure~\ref{fig:dcgf}) for obtaining DeepCrack-GF-T1\_5 curve(see section~\ref{subsec5.2} for details).}
	\label{fig:roc}
\end{figure}

To plot Receiver Operating Characteristics (ROC) (fig.~\ref{fig:rocall}) we have used sklearn [21] library and the Precision-Recall (PR) (fig.~\ref{fig:prc}) curve is plotted manually using thresholds ranging from 0 to 0.99 with a step size of 0.01. The area under curve (AUC) for DeepCrack-BN is more than other techniques (see captions for more details).

In the PR-curve plots, the curve for DeepCrack-GF is plotted by using a threshold1 (see figure~\ref{fig:dcgf}) value of $0.31$ that results in the highest F-score for the DeepCrack-BN approach (see table ~\ref{table:deepcrack_best}). For a better comparison, we have also plotted  the curve for DeepCrack-GF is plotted by using a threshold1 value of $0.5$. This plot is labelled as DeepCrack-GF-T1\_5.

The ROC curve shown in figure~\ref{fig:rocall} suggests that our models generate less false positives consistently. Our false positive rate changes marginally with different thresholds. This suggests that all the crack probabilities generated by our models are strong candidates for crack pixels (can also be validated from the results shown in tables~\ref{table:F1_best_kaggle} and~\ref{table:IOU_best_kaggle} wherein the highest F-scores are obtained around a threshold of $0.0$).

\subsection{Quantitative Analysis}\label{subsec5.3}

Tables~\ref{table:deepcrack_standard} and~\ref{table:deepcrack_best} show the comparison of our techniques with DeepCrack-BN and DeepCrack-GF on the DeepCrack dataset. The proposed truecrack approach has the highest F-score of 85.1$\%$ when thresholds of 0.5 were used on the predictions. The proposed swcrack approach achieves the highest recall of 88$\%$ whereas the allcrack approach achieves the highest precision of 94.3$\%$. See captions for more details on the performance of different approaches using the thresholds resulting the best F-scores.
Tables~\ref{table:F1_kaggle} and~\ref{table:IOU_kaggle} show the comparison of our techniques with DeepCrack-BN and DeepCrack-GF on blind datasets(see section~\ref{blinddataset}). It can be observed that the proposed swcrack model F-score is better than all the other methods on 5 datasets namely, Rissbilder, Sylvie Chambon [22], Volker, GAPs [23] and Eugen Muller. See captions of tables~\ref{table:F1_kaggle},~\ref{table:IOU_kaggle},~\ref{table:F1_best_kaggle} and~\ref{table:IOU_best_kaggle}.

\section{Conclusion}\label{sec6}

Image segmentation is one of the commonly used supervised techniques for crack detection that require labelling of each crack pixel by human experts which is expensive and time consuming. Figures~(\ref{fig:deepcrackinfbest1}-\ref{fig:volkerstandard2}) and tables~(\ref{table:deepcrack_best}-\ref{table:IOU_best_kaggle}) qualitatively and quantitatively show that our results are comparable to the state of the art techniques even after reducing the training data significantly. Our true image selection approach discussed in section~\ref{trueimageselection} has many fold advantages (a) helps in reducing the number of images to be annotated for supervised machine learning, (b) reducing training cost and time, (c) reducing carbon consumption due to less resource utilization [24]. Using knowledge based augmentation, we have further improved our model performance significantly as discussed in section~\ref{rds}.

\section{Acknowledgments}\label{sec7}

The authors would like to thank GE Research for providing the resources to enable us to do this work. 








\printcredits

\bibliographystyle{cas-model2-names}
\clearpage
\bibliography{}
\begin{enumerate}
	
	\item[{[1]}]\label{canny}
	Canny, John. "A computational approach to edge detection." IEEE Transactions on pattern analysis and machine intelligence 6 (1986): 679-698.
	
	\item[{[2]}]\label{canny_improved_crack}
	Zhao, Huili, Guofeng Qin, and Xingjian Wang. "Improvement of canny algorithm based on pavement edge detection." 2010 3rd international congress on image and signal processing. Vol. 2. IEEE, 2010.
	
	\item[{[3]}]\label{active_contours}
	Li, Shuai, Yang Cao, and Hubo Cai. "Automatic pavement-crack detection and segmentation based on steerable matched filtering and an active contour model." Journal of Computing in Civil Engineering 31.5 (2017): 04017045.
	
	\item[{[4]}]\label{random_forest}
	Chen, Jieh-Haur, et al. "A self organizing map optimization based image recognition and processing model for bridge crack inspection." Automation in Construction 73 (2017): 58-66.
	
	\item[{[5]}]\label{svm_crack}
	Hoang, Nhat-Duc, Quoc-Lam Nguyen, and Dieu Tien Bui. "Image processing–based classification of asphalt pavement cracks using support vector machine optimized by artificial bee colony." Journal of Computing in Civil Engineering 32.5 (2018): 04018037.
	
	\item[{[6]}]\label{ml_drawbacks}
	Hamishebahar, Younes, et al. "A comprehensive review of deep learning-based crack detection approaches." Applied Sciences 12.3 (2022): 1374.
	
	\item[{[7]}]\label{deepcrack}
	Liu, Yahui, et al. "DeepCrack: A deep hierarchical feature learning architecture for crack segmentation." Neurocomputing 338 (2019): 139-153.
	
	\item[{[8]}]\label{deepseg}
	"Deep Segmentor". https://github.com/yhlleo/DeepSegmentor, accessed 2021
	
	\item[{[9]}]\label{crackGAN}
	Zhang, Kaige, Yingtao Zhang, and Heng-Da Cheng. "CrackGAN: Pavement crack detection using partially accurate ground truths based on generative adversarial learning." IEEE Transactions on Intelligent Transportation Systems 22.2 (2020): 1306-1319.
	
	\item[{[10]}]\label{kagglecracksegmentationdataset}
	'Kaggle Crack Segmentation Dataset',\\ https://www.kaggle.com/datasets/lakshaymiddha/crack-segmentation-dataset, 
	accessed 2020
	
	\item[{[11]}]\label{unet}
	Ronneberger, Olaf, Philipp Fischer, and Thomas Brox. "U-net: Convolutional networks for biomedical image segmentation." Medical Image Computing and Computer-Assisted Intervention–MICCAI 2015: 18th International Conference, Munich, Germany, October 5-9, 2015, Proceedings, Part III 18. Springer International Publishing, 2015.
	
	\item[{[12]}]\label{efficienet}
	Tan, Mingxing, and Quoc Le. "Efficientnet: Rethinking model scaling for convolutional neural networks." International conference on machine learning. PMLR, 2019.
	
	\item[{[13]}]\label{imagenet}
	Deng, Jia, et al. "Imagenet: A large-scale hierarchical image database." 2009 IEEE conference on computer vision and pattern recognition. Ieee, 2009.
	
	\item[{[14]}]\label{matplotlib}
	Hunter, John D. "Matplotlib: A 2D graphics environment." Computing in science \& engineering 9.03 (2007): 90-95.
	
	\item[{[15]}]\label{coredeep}
	Pandey, R. K., \& Achara, A. (2022). CoreDeep: Improving Crack Detection Algorithms Using Width Stochasticity. arXiv preprint arXiv:2209.04648.
	
	\item[{[16]}]\label{opencv}
	Bradski, Gary. "The openCV library." Dr. Dobb's Journal: Software Tools for the Professional Programmer 25.11 (2000): 120-123.
	
	\item[{[17]}]\label{focalloss}
	Lin, Tsung-Yi, et al. "Focal loss for dense object detection." Proceedings of the IEEE international conference on computer vision. 2017.
	
	\item[{[18]}]\label{diceloss}
	Sudre, C. H., Li, W., Vercauteren, T., Ourselin, S., \& Jorge Cardoso, M. (2017). Generalised dice overlap as a deep learning loss function for highly unbalanced segmentations. In Deep learning in medical image analysis and multimodal learning for clinical decision support (pp. 240-248). Springer, Cham.
	
	\item[{[19]}]\label{albumentations}
	Buslaev, Alexander, et al. "Albumentations: fast and flexible image augmentations." Information 11.2 (2020): 125.
	
	\item[{[20]}]\label{keraslib}
	Chollet, F., \& others.  "Keras". GitHub. Retrieved from https://github.com/fchollet/keras, accessed 2015
	
	\item[{[21]}]\label{sklearn}
	Pedregosa, Fabian, et al. "Scikit-learn: Machine Learn-ing in Python, Journal of Machine Learning Re-search, 12." (2011): 2825.
	
	\item[{[22]}]\label{Amhaz2016}
	Amhaz, Rabih, et al. "Automatic crack detection on two-dimensional pavement images: An algorithm based on minimal path selection." IEEE Transactions on Intelligent Transportation Systems 17.10 (2016): 2718-2729.
	
	\item[{[23]}]\label{eisenbach2017how}
	Eisenbach, Markus, et al. "How to get pavement distress detection ready for deep learning? A systematic approach." 2017 international joint conference on neural networks (IJCNN). IEEE, 2017.
	
	\item[{[24]}]\label{carbon}
	Patterson, D., Gonzalez, J., Le, Q., Liang, C., Munguia, L.M., Rothchild, D., So, D., Texier, M. \& Dean, J. (2021). Carbon emissions and large neural network training. arXiv preprint arXiv:2104.10350.
	
\end{enumerate}

\section{Appendices}\label{sec9}

The tables below demonstrate the detailed quantitative results of the study(discussed in section~\ref{subsec5.3}). See captions of the respective tables for more details.

\begin{table*}[h]
	\centering
	\begin{subtable}[t]{\textwidth}
		\begin{tabular}{p{2.8cm}p{1.75cm}p{1.75cm}p{1.75cm}p{1.75cm}p{1.75cm}p{1.75cm}p{1.75cm}}
			\hline
			{} &  	T &     G &     C &   mIoU & P &     R &     F \\
			\hline
			DeepCrack-GF &       0.48 &  0.989 &  0.928	&  0.878 &  0.879 &  \textbf{0.858} &  \textbf{0.868} \\
			DeepCrack-BN &       0.31 &  0.987 &  0.92	&  0.864 &  0.858 &  0.846 &  0.852 \\
			allcrack     &       0.01 &  0.988 &  0.913 &  0.866 &  0.879 &  0.831 &  0.854 \\
			truecrack    &       0.03 &  0.988 &  0.920 &  0.874 &  \textbf{0.883} &  0.845 &  0.864 \\
			swcrack      &       0.99 &  0.984 &  0.898 &  0.833 &  0.820 &  0.804 &  0.812 \\
			slcrack      &       0.01 &  0.987 &  0.901 &  0.854 &  0.873 &  0.807 &  0.839 \\
			sscrack      &       0.25 &  0.986 &  0.905 &  0.853 &  0.858 &  0.817 &  0.837 \\
			mixcrack     &       0.92 &  0.986 &  0.905 &  0.853 &  0.858 &  0.817 &  0.837 \\
			\hline
		\end{tabular}
		\caption{DeepCrack-GF has the highest Recall and our truecrack has the highest Precision whereas DC-GF and truecrack have a comparable F-score.  we have used a threshold1 of 0.31 and threshold2 of 0.48 (see figure~\ref{fig:dcgf}) for obtaining DeepCrack-GF outputs.}
		\label{table:deepcrack_best}
	\end{subtable}
	
	\begin{subtable}[t]{\textwidth}
		\begin{tabular}{p{4.9cm}p{1.75cm}p{1.75cm}p{1.75cm}p{1.75cm}p{1.75cm}p{1.75cm}}
			\hline
			{} &      G &      C &    mIoU &      P &      R &     F \\
			\hline
			DeepCrack-GF &  0.988 &  0.889 &  0.863 &  0.930 &  0.782 &  0.850 \\
			DeepCrack-BN &  0.987 &  0.874 &  0.847 &  0.925 &  0.751 &  0.829 \\
			allcrack     &  0.987 &  0.873 &  0.851 &  \textbf{0.943} &  0.748 &  0.834 \\
			truecrack    &  0.988 &  0.891 &  0.864 &  0.930 &  0.784 &  \textbf{0.851} \\
			swcrack      &  0.980 &  0.932 &  0.817 &  0.719 &  \textbf{0.880} &  0.791 \\
			slcrack      &  0.986 &  0.859 &  0.837 &  0.939 &  0.720 &  0.815 \\
			sscrack      &  0.986 &  0.900 &  0.853 &  0.870 &  0.806 &  0.837 \\
			mixcrack     &  0.984 &  0.906 &  0.838 &  0.818 &  0.819 &  0.819 \\
			\hline
		\end{tabular}
		\caption{Precision, Recall and F-score are better in our techiques; we have used a threshold1 and threshold2 of 0.5 (see figure~\ref{fig:dcgf}) for obtaining DeepCrack-GF outputs.}
		\label{table:deepcrack_standard}
	\end{subtable}
	\captionsetup{width=1\textwidth}
	\caption{Table (a) shows quantitative comparison of results obtained using our techniques with that of the DeepCrack approaches; these results are obtained using a threshold that maximizes the F-score and is obtained using grid search(from 0.0 to 0.99 with a step-size of 0.01) on the DeepCrack test dataset as reported in [5], whereas in table (b) we have used a fixed threshold of 0.5 (standard threshold for binary classification) to obtain all the results.}
	\label{table:deepcrack}
\end{table*}
\begin{table*}[h]
	\centering
	\begin{subtable}[b]{\textwidth}
		\resizebox{\textwidth}{!}{%
			\begin{tabular}{lrrrrrrrr}
				\hline
				{Dataset} &  DeepCrack-GF &  DeepCrack-BN &  allcrack &  truecrack &  swcrack &  slcrack &  sscrack &  mixcrack \\
				\hline
				Rissbilder   &         0.278 &         0.154 &     0.267 &      0.255 &    \textbf{0.417} &    0.304 &    0.380 &     0.411 \\
				Sylvie Chambon       &         0.134 &         0.070 &     0.086 &      0.082 &    \textbf{0.215} &    0.069 &    0.145 &     0.187 \\
				Volker       &         0.319 &         0.220 &     0.331 &      0.318 &    \textbf{0.586} &    0.326 &    0.403 &     0.492 \\
				cfd          &         0.573 &         0.392 &     0.512 &      0.455 &    0.570 &    0.409 &    \textbf{0.595} &     0.576 \\
				crack500     &         \textbf{0.593} &         0.561 &     0.450 &      0.193 &    0.460 &    0.237 &    0.376 &     0.403 \\
				GAPs      &         0.207 &         0.170 &     0.199 &      0.287 &    \textbf{0.322} &    0.188 &    0.308 &     0.281 \\
				Eugen Muller       &         0.138 &         0.067 &     0.184 &      0.122 &    \textbf{0.496} &    0.151 &    0.273 &     0.326 \\
				forest       &         \textbf{0.615} &         0.478 &     0.522 &      0.505 &    0.589 &    0.406 &    0.607 &     0.573 \\
				cracktree200 &         0.215 &         0.171 &     \textbf{0.276} &      0.240 &    0.206 &    0.226 &    0.246 &     0.166 \\
				\hline
		\end{tabular}}
		\caption{The allcrack approach has the highest F-score on cracktree200 dataset; swcrack on 5 datasets; sscrack approach on cfd dataset; overall, our techniques are better on 7 datasets and DeepCrack-GF is better on 2 datasets namely, crack500 and forest.}
		\label{table:F1_kaggle}
	\end{subtable}
	
	\begin{subtable}[b]{\textwidth}
		\resizebox{\textwidth}{!}{%
			\begin{tabular}{lrrrrrrrr}
				\hline
				{} &  DeepCrack-GF &  DeepCrack-BN &  allcrack &  truecrack &  swcrack &  slcrack &  sscrack &  mixcrack \\
				\hline
				Rissbilder   &         0.567 &         0.528 &     0.565 &      0.561 &    \textbf{0.620} &    0.577 &    0.606 &     0.618 \\
				Sylvie Chambon       &         0.498 &         0.480 &     0.487 &      0.486 &    \textbf{0.528} &    0.482 &    0.505 &     0.518 \\
				Volker       &         0.577 &         0.543 &     0.582 &      0.577 &    \textbf{0.694} &    0.580 &    0.610 &     0.648 \\
				cfd          &         0.696 &         0.616 &     0.667 &      0.642 &    0.692 &    0.623 &    \textbf{0.707} &     0.696 \\
				crack500     &         \textbf{0.687} &         0.673 &     0.622 &      0.526 &    0.625 &    0.540 &    0.591 &     0.602 \\
				GAPs      &         0.547 &         0.538 &     0.548 &      0.577 &    \textbf{0.588} &    0.544 &    0.583 &     0.575 \\
				Eugen Muller       &         0.513 &         0.493 &     0.530 &      0.511 &    \textbf{0.648} &    0.519 &    0.559 &     0.578 \\
				forest       &         \textbf{0.717} &         0.652 &     0.672 &      0.664 &    0.702 &    0.622 &    0.713 &     0.695 \\
				cracktree200 &         0.555 &         0.543 &     \textbf{0.577} &      0.565 &    0.552 &    0.561 &    0.565 &     0.536 \\
				\hline
		\end{tabular}}
		\caption{The allcrack approach has the highest mIoU on cracktree200 dataset; swcrack approach on 5 datasets; sscrack approach on cfd dataset; overall, our techniques are better on 7 datasets and DeepCrack-GF is better on 2 datasets namely, crack500 and forest.}
		\label{table:IOU_kaggle}
	\end{subtable}
	\caption{Tables (a) and (b) show the F1 and mIoU comparision of our techniques with that of DeepCrack approaches respectively. The scores have been calculated at a threshold of 0.5 over multiple blind datasets (see~\ref{blinddataset}).}
	\label{table:standard}
\end{table*}

\begin{table*}[ht]
	\centering
	\begin{subtable}[t]{\textwidth}
		\resizebox{\textwidth}{!}{%
			\begin{tabular}{lllllllll}
				\hline
				{} &          DeepCrack-GF &   DeepCrack-BN &       allcrack &      truecrack &        swcrack &        slcrack &        sscrack &       mixcrack \\
				\hline
				Rissbilder   &  (0.02, 0.431) &  (0.06, 0.449) &   (0.0, 0.422) &    (0.0, 0.42) &   (0.0, 0.463) &   (0.0, 0.402) &   \textbf{(0.0, 0.487)} &  (0.01, 0.471) \\
				Sylvie Chambon       &   \textbf{(0.0, 0.372)} &  (0.03, 0.335) &   (0.0, 0.195) &   (0.0, 0.183) &   (0.0, 0.349) &   (0.0, 0.183) &    (0.0, 0.27) &   (0.0, 0.363) \\
				Volker       &  (0.01, 0.542) &  (0.04, 0.549) &   (0.0, 0.506) &   (0.0, 0.476) &   \textbf{(0.0, 0.695)} &   (0.0, 0.476) &   (0.0, 0.549) &   (0.0, 0.637) \\
				cfd          &   (0.2, 0.618) &  (0.18, 0.639) &   \textbf{(0.0, 0.661)} &   (0.0, 0.611) &  (0.92, 0.576) &    (0.0, 0.55) &  (0.01, 0.619) &   (0.2, 0.577) \\
				crack500     &  \textbf{(0.26, 0.601)} &  (0.25, 0.597) &   (0.0, 0.563) &   (0.0, 0.316) &   (0.0, 0.541) &   (0.0, 0.384) &   (0.0, 0.482) &   (0.0, 0.503) \\
				GAPs      &  (0.14, 0.227) &  (0.18, 0.225) &   (0.0, 0.338) &   \textbf{(0.0, 0.345)} &  (0.02, 0.332) &  (0.21, 0.189) &   (0.02, 0.33) &  (0.01, 0.292) \\
				Eugen Muller        &   (0.0, 0.376) &  (0.03, 0.401) &   (0.0, 0.399) &   (0.0, 0.285) &   \textbf{(0.0, 0.635)} &   (0.0, 0.317) &   (0.0, 0.418) &   (0.0, 0.537) \\
				forest       &  (0.26, 0.641) &  (0.21, 0.655) &   \textbf{(0.0, 0.667)} &   (0.0, 0.619) &   (0.66, 0.59) &   (0.0, 0.568) &   \textbf{(0.01, 0.64)} &   (0.07, 0.58) \\
				cracktree200 &   (0.65, 0.22) &  (0.34, 0.207) &  \textbf{(0.05, 0.287)} &  (0.06, 0.256) &  (0.98, 0.222) &  (0.04, 0.239) &   (0.98, 0.27) &  (0.99, 0.223) \\
				\hline
		\end{tabular}}
		\caption{Each tuple has the form (F-score, Threshold). The allcrack approach achieves the highest F-score on 3 datasets; truecrack on GAPs; swcrack on Volker and Eugen Muller; sscrack on Rissbilder; overall, our techniques are better on 7 datasets and DeepCrack-GF (column2 thresholds are threshold1 and column1 thresholds are threshold2, see Figure~\ref{fig:dcgf}) is better on crack500 and Sylvie Chambon.}
		\label{table:F1_best_kaggle}
	\end{subtable}
	\begin{subtable}[t]{\textwidth}
		\resizebox{\textwidth}{!}{%
			\begin{tabular}{lllllllll}
				\hline
				{} &          DeepCrack-GF &   DeepCrack-BN &      allcrack &     truecrack &        swcrack &        slcrack &        sscrack &       mixcrack \\
				\hline
				Rissbilder   &  (0.02, 0.621) &  (0.06, 0.628) &  (0.0, 0.623) &  (0.0, 0.621) &   (0.0, 0.637) &   (0.0, 0.611) &   \textbf{(0.0, 0.649)} &  (0.01, 0.641) \\
				Sylvie Chambon       &   (0.0, 0.577) &  (0.03, 0.555) &   (0.0, 0.52) &  (0.0, 0.517) &   (0.0, 0.575) &   (0.0, 0.517) &   (0.0, 0.547) &    \textbf{(0.0, 0.58)} \\
				Volker       &   (0.01, 0.67) &  (0.04, 0.672) &  (0.0, 0.655) &  (0.0, 0.641) &   \textbf{(0.0, 0.755)} &   (0.0, 0.641) &   (0.0, 0.675) &   (0.0, 0.721) \\
				cfd          &   (0.2, 0.718) &  (0.18, 0.729) &  \textbf{(0.0, 0.742)} &  (0.0, 0.714) &  (0.92, 0.696) &   (0.0, 0.684) &  (0.01, 0.718) &   (0.2, 0.696) \\
				crack500     &  \textbf{(0.26, 0.689)} &  (0.25, 0.688) &  (0.0, 0.674) &  (0.0, 0.566) &   (0.0, 0.659) &   (0.0, 0.593) &   (0.0, 0.635) &   (0.0, 0.643) \\
				GAPs      &   (0.14, 0.55) &  (0.18, 0.548) &  (0.0, 0.594) &  \textbf{(0.0, 0.595)} &  (0.02, 0.591) &  (0.21, 0.544) &   (0.02, 0.59) &  (0.01, 0.578) \\
				Eugen Muller        &    (0.0, 0.59) &  (0.03, 0.598) &  (0.0, 0.606) &  (0.0, 0.563) &   \textbf{(0.0, 0.718)} &   (0.0, 0.575) &   (0.0, 0.613) &   (0.0, 0.667) \\
				forest       &   (0.26, 0.73) &  (0.21, 0.738) &  \textbf{(0.0, 0.745)} &  (0.0, 0.719) &  (0.66, 0.703) &   (0.0, 0.693) &   (0.01, 0.73) &  (0.07, 0.698) \\
				cracktree200 &  (0.65, 0.557) &  (0.34, 0.552) &  \textbf{(0.05, 0.58)} &  (0.06, 0.57) &  (0.98, 0.559) &  (0.04, 0.564) &  (0.98, 0.575) &  (0.99, 0.558) \\
				\hline
		\end{tabular}}
		\caption{The allcrack approach has the highest mIoU on 3 datasets; truecrack on GAPs; swcrack on Volker and Eugen Muller; sscrack on Rissbilder; mixcrack on Sylvie Chambon; overall, our techniques are better on 8 datasets and DeepCrack-GF (column2 thresholds are threshold1 and column1 thresholds are threshold2, see Figure~\ref{fig:dcgf}) is better on crack500.}
		\label{table:IOU_best_kaggle}
	\end{subtable}
	\caption{Tables (a) and (b) show the F-score and mIoU comparision of our techniques with that of DeepCrack approaches respectively. The scores have been calculated at the best thresholds (grid search from 0.0 to 0.99 at a step of 0.01) over multiple blind datasets (see~\ref{blinddataset}).}
	\label{table:overall}
\end{table*}

\end{document}